\documentclass[fleqn,10pt]{wlscirep}
\usepackage[utf8]{inputenc}
\usepackage[T1]{fontenc}
\usepackage{colortbl}
\usepackage{xcolor}
\usepackage{bm}
\usepackage{multirow}
\usepackage{url}

\title{OpenUAS: Embeddings of Cities in Japan with Anchor Data for Cross-city Analysis of Area Usage Patterns}

\author[1]{Naoki Tamura}
\author[1]{Kazuyuki Shoji}
\author[1]{Shin Katayama}
\author[1]{Kenta Urano}
\author[1]{Takuro Yonezawa}
\author[1,2]{Nobuo Kawaguchi}
\affil[1]{Graduate School of Engineering, Nagoya University}
\affil[2]{Institutes of Innovation for Future Society, Nagoya University}

\begin{abstract}

We publicly release OpenUAS, a dataset of area embeddings based on urban usage patterns, including embeddings for over 1.3 million 50-meter square meshes covering a total area of 3,300 square kilometers. This dataset is valuable for analyzing area functions in fields such as market analysis, urban planning, transportation infrastructure, and infection prediction. It captures the characteristics of each area in the city, such as office districts and residential areas, by employing an area embedding technique that utilizes location information typically obtained by GPS. Numerous area embedding techniques have been proposed, and while the public release of such embedding datasets is technically feasible, it has not been realized. One reason for this is that previous methods could not embed areas from different cities and periods into the same embedding space without sharing raw location data. We address this issue by developing an anchoring method that establishes anchors within a shared embedding space. We publicly release this anchor dataset along with area embedding datasets from several periods in eight major Japanese cities.

\end{abstract}
\begin{document}

\flushbottom
\maketitle

\thispagestyle{empty}

\section*{Background \& Summary}

Understanding how people use different sections and areas of a city is crucial across a wide range of fields\cite{anomaly, dynamics, poi_recommend1, poi_recommend2, poi_recommend3, human_mob_pred1, human_mob_pred2, human_mob_pred3, urbanFunctionCluster}. This includes commercial applications such as market analysis\cite{poi2vec} and store location strategies\cite{store_location}, as well as urban planning\cite{urban_function1, urban_function2, urban_function3}, traffic infrastructure development\cite{traffic_sim}, and infection prediction\cite{infection}. For these applications, research has focused on analyzing area functions based on extensive location data, such as GPS (Global Positioning System)\cite{survey1, survey2, survey3}. Efforts to publicly release GPS-based human mobility datasets have been actively pursued\cite{tdrive, geolife, nyc_tlc_2023, crawdad_roma, crawdad_SF}. Especially due to the COVID-19 pandemic, the release of human flow datasets for analyzing the spread of infections has advanced\cite{covid19_flow1, covid19_flow2, covid19_mob_change}. However, many of these datasets are limited to taxi movements or inter-regional OD data, and have low spatiotemporal resolution, making it difficult to analyze area usage patterns on a metropolitan scale. To overcome these challenges, Yabe et al.\cite{YJmob100K} publicly released the YJMob100k dataset, a city-scale longitudinal human mobility dataset. However, to preserve privacy, they are only provided at a coarse spatial resolution of about 500m mesh, making them unsuitable for detailed spatial analysis in urban functions. 

On the other hand, area embedding techniques have been proposed, aggregating features derived from location information by area and mapping them onto a vector space\cite{areavec1, areavec2, areavec3, poi2vec2, hier_emb}. Such embeddings allow for advanced area analysis by aggregating information for each area, which eliminates personal data and enables searching and clustering of areas with similar functions and usage patterns. In this paper, we apply the embedding method called Area2Vec proposed by Shoji et al\cite{area2vec}. This method embeds areas based on stay information extracted from GPS data, resulting in embeddings called UAS (Usage of Area with Stay information) that reflect how people use the areas. The UAS can capture not only the static functions of urban areas but also the dynamic changes in area functions by tracking shifts in area usage patterns due to environmental changes(urban development, seasonal changes, pandemics). By adopting UAS as the public area embedding dataset, it becomes possible to analyze changes in area functions over multiple periods, in addition to comparing area functions between cities.

Releasing area embedding datasets provides valuable information for analyzing area functions to users who cannot directly access raw GPS datasets. Despite the demand for the public release of area embedding datasets, existing methods have not released area embedding datasets. One reason is that area embeddings, learned independently in each dataset, are unique representations. They embed the relative similarities between areas within a dataset into an embedding space. This uniqueness poses a challenge: areas from different datasets, such as cities or periods, cannot be directly compared. For example, embeddings learned independently for areas in New York and Tokyo cannot be compared due to the lack of a common embedding space. Similarly, embeddings of the same city from different periods, such as pre-COVID and post-COVID periods, are incomparable. This issue can be addressed by combining datasets of New York and Tokyo or by training new translation models\cite{city2city, embacrosscity}. However, if the location data for each city is owned by different organizations, sharing the actual location data becomes difficult, making these methods impractical. Therefore, we cannot compare them with existing publicly available embeddings, even if new embeddings are generated for a different city (or a different period) using their datasets. 

To address this challenge, we have developed a method that anchors the embedding space by establishing several standard vectors, such as office anchor or residential anchor, within it and fixing their positions. Initially, our method samples anchor data by removing geographical information from our GPS data to ensure it does not contain any private information. Then, we fix the embeddings for the anchor data. By learning the embeddings for each area under those conditions, we can acquire area embeddings based on the fixed anchor positions. This allows for the embedding of areas in a common space and enables comparison between areas across datasets without the need to share the datasets themselves. Applying this method, we publicly release OpenUAS, a dataset of area embeddings based on urban usage patterns for eight major cities in Japan based on extensive location information(covering populations ranging from 5\% to 10\% in scale), as well as data for several periods from 2019 to 2023 for the city of Nagoya. Additionally, OpenUAS includes anchor data, allowing data holders to compare their individual data with the embeddings of areas in various Japanese cities. This dataset is intended to be of broad utility for researchers and data analysts, supporting a wide range of applications, including land use analysis, inter-city comparisons, and area marketing in urban environments.

The contributions of this paper are as follows:
\begin{itemize}
    \item Development of an anchoring method enabling embedding and analyzing areas across datasets.
    \item The public release of area embedding data for eight major cities in Japan with anchor data based on extensive location information. (\url{https://zenodo.org/records/13141800})
    \item Development and release of tools and codes for effective use of area embedding data. (\url{https://github.com/UCLabNU/OpenUAS/})
\end{itemize}

The next section will detail the area embedding and anchoring methods. In Data Records, we will explain the specifics of the area embedding data, and in Validation, we will assess the effectiveness and anonymity of the embedding representations and the anchoring method. Finally, in Usage Notes, we will discuss effective ways to utilize area embedding data, including the anchoring method, and provide relevant code.

\section*{Methods}
\subsection*{Data sources and summary of our dataset.}
First, we will explain the GPS dataset we use for area embeddings and the dataset released in this paper. We release area embeddings using Area2Vec method and the anchoring method described later. Area2Vec embeds areas based on stay information in each area. We first obtained GPS data, which includes anonymized user IDs, latitude, longitude, and timestamps, from Blogwatcher Inc. This data is obtained with the use of multiple applications on mobile devices with user consent. To extract stay information from the GPS data, we used the stay estimation method by Iwata et al\cite{stay_est}. The resulting stay dataset includes columns for latitude, longitude, day of the week, time, and stay duration for each stay, where the day of the week and time indicate the start of the stay. 

\begin{table}[h]
\small
\begin{tabular}{|c|c|c|c|c|>{\columncolor[gray]{0.9}}c|}
\hline
\textbf{\begin{tabular}{c}Dataset \\Name\end{tabular}} & \textbf{City} & \textbf{Period} & \textbf{\begin{tabular}{c}\# of 250m mesh \\embeddings\end{tabular}} & \textbf{\begin{tabular}{c}\# of 50m mesh \\embeddings\end{tabular}} & \textbf{\begin{tabular}{c}
\# of source stay data records \\ (not for public release)
\end{tabular}}\\
\hline
 $D_{Tokyo}$ & Tokyo & 2021/04/01-2021/04/30 & 4,277 &194,155&41,086,448 \\
 $D_{Osaka}$ & Osaka & 2021/04/01-2021/04/30 & 2,229 &61,214&10,235,553 \\
 $D_{Sapporo}$ & Sapporo & 2021/04/01-2021/04/30 & 5,374 &45,679&5,698,663 \\
 $D_{Fukuoka}$ & Fukuoka & 2021/04/01-2021/04/30 & 3,133 &35,779&4,903,561 \\ 
 $D_{Sendai}$ & Sendai & 2021/04/01-2021/04/30 & 3,942 &24,431&3,335,943 \\
 $D_{Hiroshima}$ & Hiroshima & 2021/04/01-2021/04/30 & 3,795 &23,817&3,203,044 \\ 
 $D_{Kyoto}$ & Kyoto & 2021/04/01-2021/04/30 & 2,947 &25,550&2,526,688 \\
\hline
\end{tabular}
\caption{Area embeddings for 8 Japanese cities.}
\label{loc_data_city}
\end{table}

\begin{table}[h]
\centering
\small
\begin{tabular}{|c|c|c|c|c|>{\columncolor[gray]{0.9}}c|}
\hline
\textbf{\begin{tabular}{c}Dataset \\Name\end{tabular}} & \textbf{City} & \textbf{Period} & \textbf{\begin{tabular}{c}\# of 250m mesh \\embeddings\end{tabular}} & \textbf{\begin{tabular}{c}\# of 50m mesh \\embeddings\end{tabular}} & \textbf{\begin{tabular}{c}
\# of source stay data records \\ (not for public release)
\end{tabular}}\\
\hline
$D_{2019}$ & \multirow{16}{*}{Nagoya} & 2019/04/01-2019/04/30 & 4,070 &77,383&11,019,245 \\
$D_{2020}$ & &2020/04/01-2020/04/30 & 4,441 &67,492& 8,306,364  \\
$D_{2021}$ & &2021/04/01-2021/04/30 & 4,682 &62,311& 7,683,285  \\
$D_{2022}$& & 2022/04/01-2022/04/30 & 4,651 &47,835& 5,514,662  \\
$D_{2023}$& & 2023/04/01-2023/04/30 & 4,692 &34,790& 3,381,312  \\
$D_{Jan}$& & 2021/01/01-2023/01/31 & 4,544 &62,503&7,874,868 \\
$D_{Feb}$& & 2021/02/01-2023/02/28 & 4,558 &61,865&7,506,891  \\
$D_{Mar}$& & 2021/03/01-2023/03/31 & 4,438 &66,159&8,556,642  \\
$D_{May}$& & 2021/05/01-2023/05/31 & 4,610 &57,986&6,916,965  \\
$D_{Jun}$& & 2021/06/01-2023/06/30 & 4,628 &55,607&6,690,968  \\
$D_{Jul}$& & 2021/07/01-2023/07/31 & 4,608 &54,253& 6,675,384  \\
$D_{Aug}$& & 2021/08/01-2023/08/31 & 4,650 &50,955& 6,081,658  \\
$D_{Sep}$& & 2021/09/01-2023/09/30 & 4,672 &51,218& 5,978,458  \\
$D_{Oct}$& & 2021/10/01-2023/10/31 & 4,624 &54,317& 6,542,341  \\
$D_{Nov}$& & 2021/11/01-2023/11/30 & 4,648 &52,551& 6,353,190  \\
$D_{Dec}$& & 2021/12/01-2023/12/31 & 4,636 &52,128& 6,433,529  \\
\hline
\end{tabular}
\caption{Area embeddings for multiple time periods in Nagoya.}
\label{loc_data_period}
\end{table}

We use stay datasets from multiple regions in Japan and release the corresponding area embeddings. The target regions are major Japanese cities: Tokyo, Osaka, Nagoya, Sapporo, Fukuoka, Sendai, Hiroshima, and Kyoto. For each city, we acquired data for one month and used this information to embed each area. Table \ref{loc_data_city} shows the number of area embeddings for each city and the number of records in the underlying data (which is not publicly available).  Additionally, as the functions of areas change due to environmental changes, longer-term data can potentially analyze these changing trends. Consequently, we focused solely on Nagoya, obtaining annual data for the month of April from 2019 to 2023, as well as monthly data from January to December 2021 (Table \ref{loc_data_period}). The annual data aims to analyze area function changes before and after the COVID-19 outbreak, and the monthly data focuses on analyzing monthly changes in 2021, a year marked by frequent emergency declarations due to the pandemic. Details of each dataset record are presented in Table \ref{loc_data_city} and Table \ref{loc_data_period}. The number of meshes, which will be explained in the following section, is included. These datasets will be referred to as indicated in the `Dataset Name' in Tables. 

\subsection*{Area Definition}
Area embedding involves defining the shape of each area, aggregating the features within each area, and then embedding them. Traditional methods define areas in various ways, such as points\cite{areavec2, areavec3, poi2vec2}, meshes\cite{city2city, hier_emb}, or polygons\cite{areavec1}. OpenUAS defines areas as equally gridded meshes and aggregate features for each mesh. It is desirable for each mesh to have a sufficient number of records for embedding, and from a privacy perspective, each mesh should contain a substantial number of unique users. For these reasons, we use two mesh sizes, 50m $\times$ 50m and 250m $\times$ 250m, following these steps for feature aggregation:
\begin{enumerate}
    \item Assign each record in the location dataset to the corresponding 50m mesh.
    \item Count the unique number of users in each 50m mesh and exclude meshes with 10 or fewer users.
    \item Assign the records excluded in step 2 to the corresponding 250m mesh (excluding those already counted in the 50m mesh).
    \item Count the unique number of users in each 250m mesh and exclude meshes with 10 or fewer users.
\end{enumerate}

By aggregating data in this way, each mesh will always have records from more than 10 unique users. Therefore, this approach mitigates the risk of embedding individual life patterns in meshes, such as residential areas, while ensuring each mesh has enough data size for embedding.

\subsection*{Area Embedding}
There are several variations of the Area Embedding method, which abstracts the functions of areas into vectors by embedding aggregated features into a latent space. Some researchers, like Yao et al.\cite{areavec1}, Crivellari et al.\cite{areavec2}, Zhai et al.\cite{place2vec}, and Liu et al.\cite{areavec3}, propose methods that model area proximity as embeddings based on in-out information of each area. In these methods, areas with similar in-out characteristics (many users come from the same area or many users go to) are embedded closer in the vector space. On the other hand, Shoji et al\cite{area2vec}. propose Area2Vec, which embeds areas using stay information, where areas with similar stay characteristics are embedded closely in the vector space. Stay information for each area includes details on how the area is used during specific periods. Using this information, we can obtain embeddings that capture the dynamically changing usage of each area in the city. The usage of each area reflects urban functions and greatly contributes to their analysis. We adopt the Area2Vec method for area embeddings to model urban functions. The reason for this choice is that Area2Vec uses only stay information, which is not tied to specific geographic factors, making it ideal for sharing area embeddings across regions. Additionally, by obtaining embeddings over multiple periods with Area2Vec, it is possible to analyze changes in area functions over time, not just compare area functions between cities. Furthermore, stay information is easier to interpret compared to in-out information, facilitating comparisons of area functions between different cities and periods.

\begin{figure}[t]
\centering
\includegraphics[width=\linewidth]{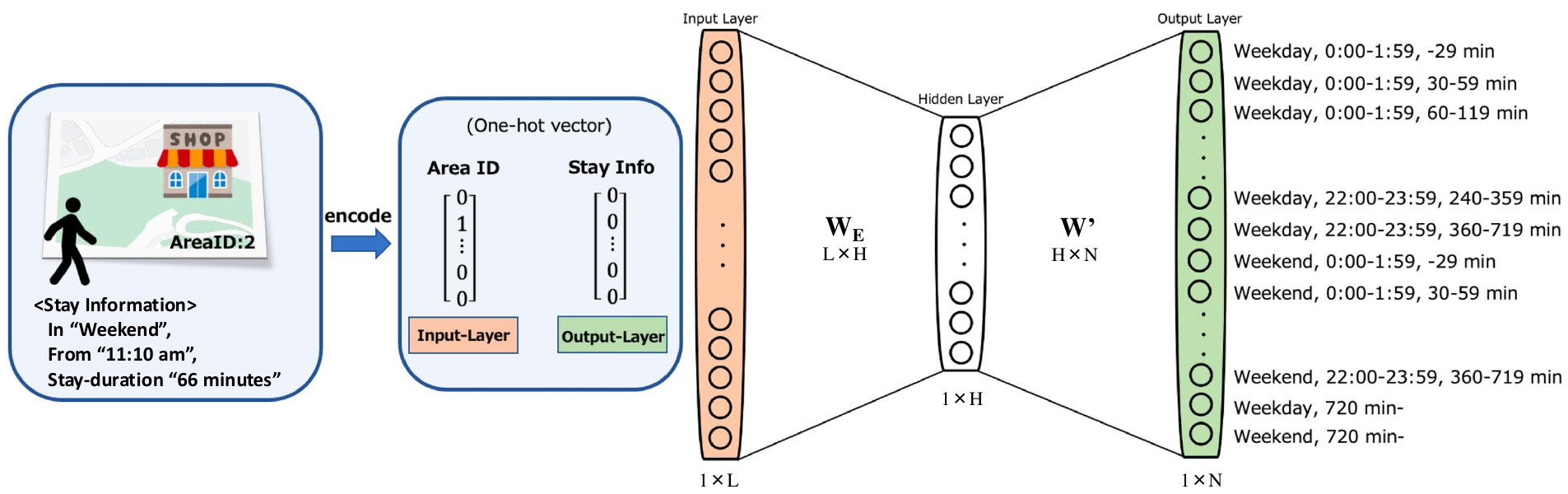}
\caption{Area2Vec architecture (Adapted from Shoji et al., 2021\cite{area2vec}).}
\label{fig:arc_of_Area2Vec}
\end{figure}

The architecture of Area2Vec is shown in Figure \ref{fig:arc_of_Area2Vec}\cite{area2vec}. The input is a one-hot vector corresponding to each area ID in the dataset, and the output is a one-hot vector corresponding to each discretized stay feature. As shown in Table \ref{table_stay_feature}, the discretization of stay features involves assigning the day of the week, arrival time(start time of each stay), and stay duration to discrete classes. This discretization results in the stay characteristics being classified into \begin{math}2*12*7=168\end{math} categories, and therefore, in this paper, \begin{math}N' = 168\end{math}. The weight of the input layer W corresponds to the Area Embedding, and in this paper, by setting $H=8$, we obtain an 8-dimensional embedding. By learning the relationships between areas and discretized stay features, areas with similar usage patterns are embedded close to each other in the vector space, while dissimilar areas are embedded farther apart. In more detail, when a one-hot vector of each area ID is inputted, the model is trained to approximate a 168-dimensional statistical vector, which represents the aggregated discretized stay characteristics of that area. Consequently, areas with similar statistical vectors will have similar vectors, resulting in the 168 dimensions being compressed into an 8-dimensional embedding. Therefore, clustering of area embeddings yields results as shown in Figure \ref{fig:cluster5}. This figure will be explained in detail in the Technical Validation section.

\begin{table}[h]
\centering
\begin{tabular}{|l|l|}
\hline
\textbf{Day of the week} & Weekday, Weekend or Holiday\\
\hline
\textbf{Arrival time} & 0:00-1:59, 2:00-3:59, ... 22:00-23:59\\
\hline
\textbf{Stay duration} & -29min, 30-59min, 60-119min, 120-239min, 240-360min, 360-720min, 720min-\\
\hline
\end{tabular}
\caption{Discretized area stay features.}
\label{table_stay_feature}
\end{table}

\subsection*{Anchoring of embedding space}
We develop an anchoring method to fix the embedding space across datasets, allowing the embedding of areas within each dataset into a common space. This enables the comparison of area embeddings across different datasets. First, we sample area data (anchor data) that includes typical areas such as residential and office districts. Then, during the embedding process, this anchor data is mixed with the dataset for training, fixing the embedding values corresponding to the Anchor. This approach fixes the latent meanings of each dimension in the embedding space across datasets. The following sections will sequentially explain the details of the anchoring method.

\begin{figure}[t]
\centering
\includegraphics[width=\linewidth]{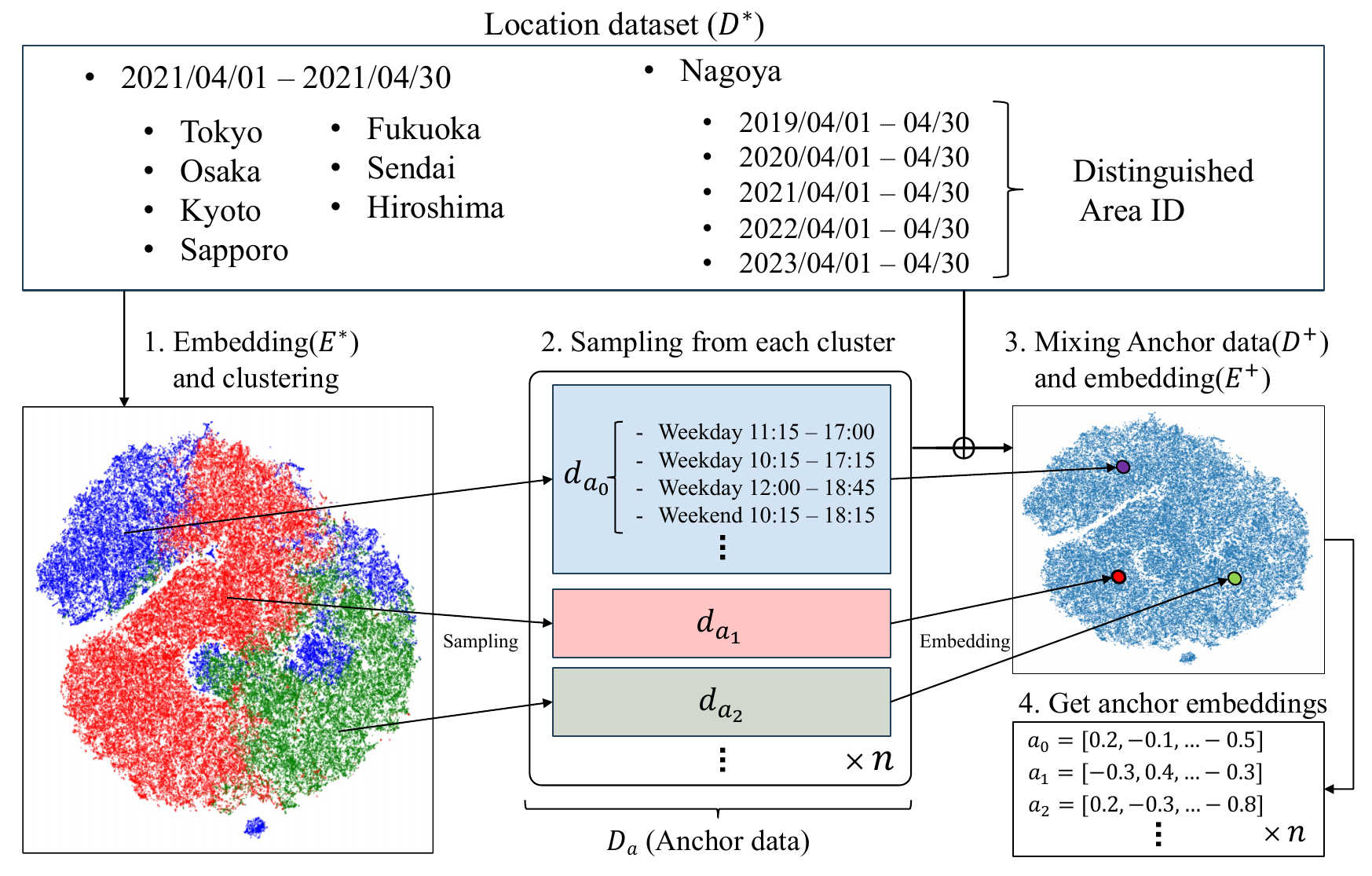}
\caption{Anchor data is sampled from location dataset. Data from multiple cities and periods are embedded and clustered. Samples from each cluster are used as anchor data, and reference values are obtained as embeddings for each anchor.}
\label{fig:gen_anchor}
\end{figure}

First, we sample a group of anchor data $D_a = \{d_{a_n}\}$, including typical area data such as residential and office districts. The overview of the sampling of anchor data is shown in Figure \ref{fig:gen_anchor}. Each anchor data $d_{a_n}$ corresponds to the $n$th anchor $a_n$, which needs to be evenly distributed in the latent space. In other words, each anchor data $d_{a_n}$ should ideally represent a different area usage pattern, and $D_a$ as a whole should encompass a diverse range of area tendencies. To sample such $D_a$, we embedded a large location dataset $D^*$ into a single latent space, followed by clustering, and used data sampled from each cluster as the respective anchor data $d_{a_n}$. $D^*$ includes areas from eight cities in Japan, with data from Nagoya covering the years 2019-2023 (treating the same area as different for each year). The resulting $D_a$ includes unique area tendencies for each of the eight cities, as well as a variety of area tendencies such as pre-COVID, during the pandemic, and post-pandemic. For the public release data, we set the number of anchors $n = 512$ and the amount of data for each anchor data at $20,000$. Optimization of the number of anchors and the amount of data is described in the appendix. In the $D_a$, timestamp and stay duration are recorded in 15-minute intervals to ensure that the stays of specific individuals cannot be extracted from $D_a$. 

\begin{table}[h]
\centering
\begin{tabular}{|c|c|c|c|c|}
\hline 
\textbf{Dataset Name} & \textbf{City} & \textbf{Period} & \textbf{\# of anchor embeddings} & \textbf{\begin{tabular}{c}
\# of source stay data records \\ 
\end{tabular}}\\
\hline
\multirow{6}{1.3cm}{$D_{a}$}& \begin{tabular}{c}
Tokyo, \\Osaka, \\
Sapporo, \\Fukuoka,\\
Sendai, \\Hiroshima,\\
Kyoto
\end{tabular} & 2021/04/01 - 2021/04/30 & \multirow{6}{3cm}{512} & \multirow{6}{3cm}{\begin{tabular}{c}10,240,000\\(512$\times$20,000)\end{tabular}}\\
\cline{2-3}
            & Nagoya & \begin{tabular}{c}
2019/04/01 - 2019/04/30 \\ 
2020/04/01 - 2020/04/30 \\
2021/04/01 - 2021/04/30 \\
2022/04/01 - 2022/04/30 \\
2023/04/01 - 2023/04/30
\end{tabular} &  &\\
\hline
\end{tabular}
\caption{Summary of the anchor dataset.}
\label{hidden_size}
\end{table}

After sampling the anchor data $D_a$, we optimize the value of the anchor embedding $a_n$ corresponding to each $d_{a_n}$ to determine where to place each anchor $a_n$ in the latent space. To do this, we prepare a dataset $D^+$, by mixing $D_a$ and $D^*$($D^+ = D_a + D^*$). By training on this $D^+$, we obtain the embeddings $E^+$. $E^+$ includes the embeddings $a_n$ for each $d_{a_n}$ which are optimized as the positions of anchors on $E^+$ that cover various trend areas. When training on a new dataset, fixing the anchors at $a_n$ allows different datasets to be embedded in a common latent space.

\begin{figure}[t]
\centering
\includegraphics[width=\linewidth]{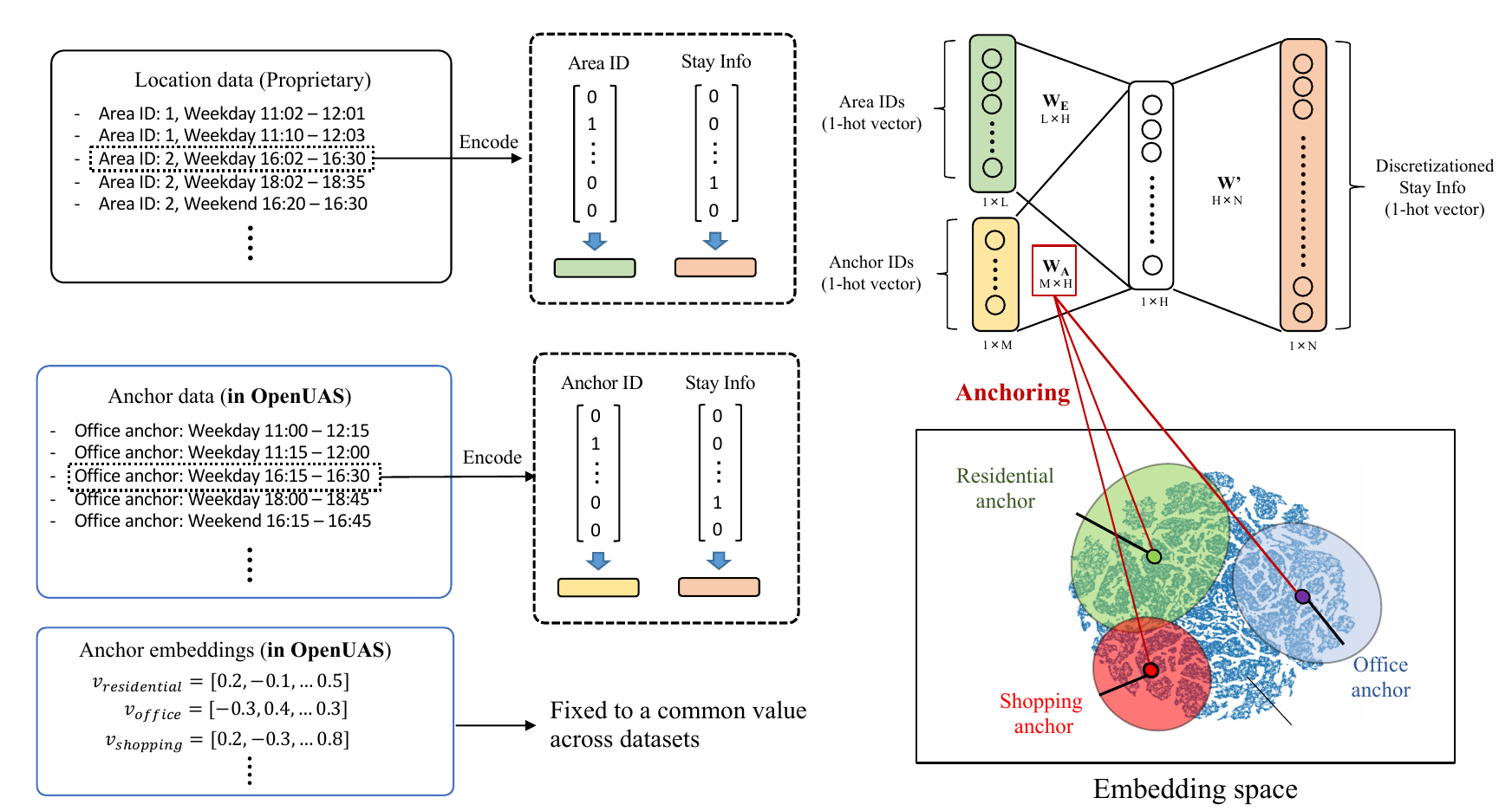}
\caption{Embedding with anchors. Proprietary data is mixed with anchor data to learn embeddings and fix the anchor embeddings to common values. Embeddings for proprietary data are aligned based on the anchors.}
\label{fig:train_anchor}
\end{figure}

Finally, we explain the method of embedding area data from different datasets using anchoring, as illustrated in Figure \ref{fig:train_anchor}. During the embedding process, we mix anchor data $D_a$ with the dataset for training and fix the values of the Anchor embeddings $W_A$ (a matrix consisting of $a_n$ values). This approach fixes the latent meanings of each dimension in the embedding space across datasets. However, if we simply implement this(corresponding to `Mixed' in Figure \ref{fig:anchor_weight} of the appendix), and the count of areas to be embedded is much larger than the anchor data, the anchoring can be ignored during learning. Therefore, we adjust the loss for the anchor during training according to Equation \ref{anchor_loss} and \ref{anchor_power}. The term $t$ is the current epoch number, $T$ refers to the total number of epochs, and $p$ represents the strength of the anchoring effect on learning, which we refer to as `Anchoring power.' As shown in Equation \ref{anchor_power}, we set the Anchoring power high in the early stages of learning ($p=\beta$ at $t/T=0.0$) and exponentially decrease it towards the latter stages ($p=\alpha$ at $t/T=1.0$). This way, in the early stages of learning, the embedding of each area learns its relationship with the anchors, while in the later stages, the relationships among individual areas are learned. We set $\beta=1.0$ and $\alpha=0.3$. The optimization of these parameters is described in the appendix. This enables other users to embed areas from other location datasets into the same representation space through the anchor data. 

\begin{equation}
   Loss = (1-p)Loss_{data} + pLoss_{anchor}
   \label{anchor_loss}
\end{equation}

\begin{equation}
   p = exp(\frac{t}{T}(\log\alpha - \log\beta) + \log\beta)
   \label{anchor_power}
\end{equation}

\section*{Data Records}
The OpenUAS can be accessed at zenodo(\url{https://zenodo.org/records/13141800}). Embedding data for each dataset is aggregated every 50m or 250m mesh and stored as vector information. The total number of meshes is 1,418,827. The data columns include geocode, Latitude, Longitude, Geometry, vector, cluster5, cluster10, and cluster20. The geocode is an identifier assigned to each standard mesh, with 250m meshes having 10-digit and 50m meshes having 12-digit integer values. Latitude and Longitude indicate the center coordinates of each mesh. Geometry and Vector are lists of floating-point numbers, with the former representing the mesh polygon and the latter the vector representation of the area embedding. cluster5, cluster10, and cluster20 are the numbers assigned when the area embeddings are clustered into those respective numbers. Graphs showing the area usage patterns of each cluster in these clusterings are also included.

Additionally, we released the anchor embeddings and anchor dataset to enable users to embed their own data into the same latent space using the anchoring method. The Anchor Embeddings are defined as $512\times8$ dimensional tensors in PyTorch and are stored as .pt files. The anchor data is a $512\times20,000$ record .csv file containing anchor\_id, arrival\_time, and stay\_time. All codes, including those for embedding and area analysis, are released at the GitHub repository(\url{https://github.com/UCLabNU/OpenUAS/}). Detailed explanations on how to use them are provided in the Usage Note and README.

\begin{figure}[t]
\centering
\includegraphics[width=\linewidth]{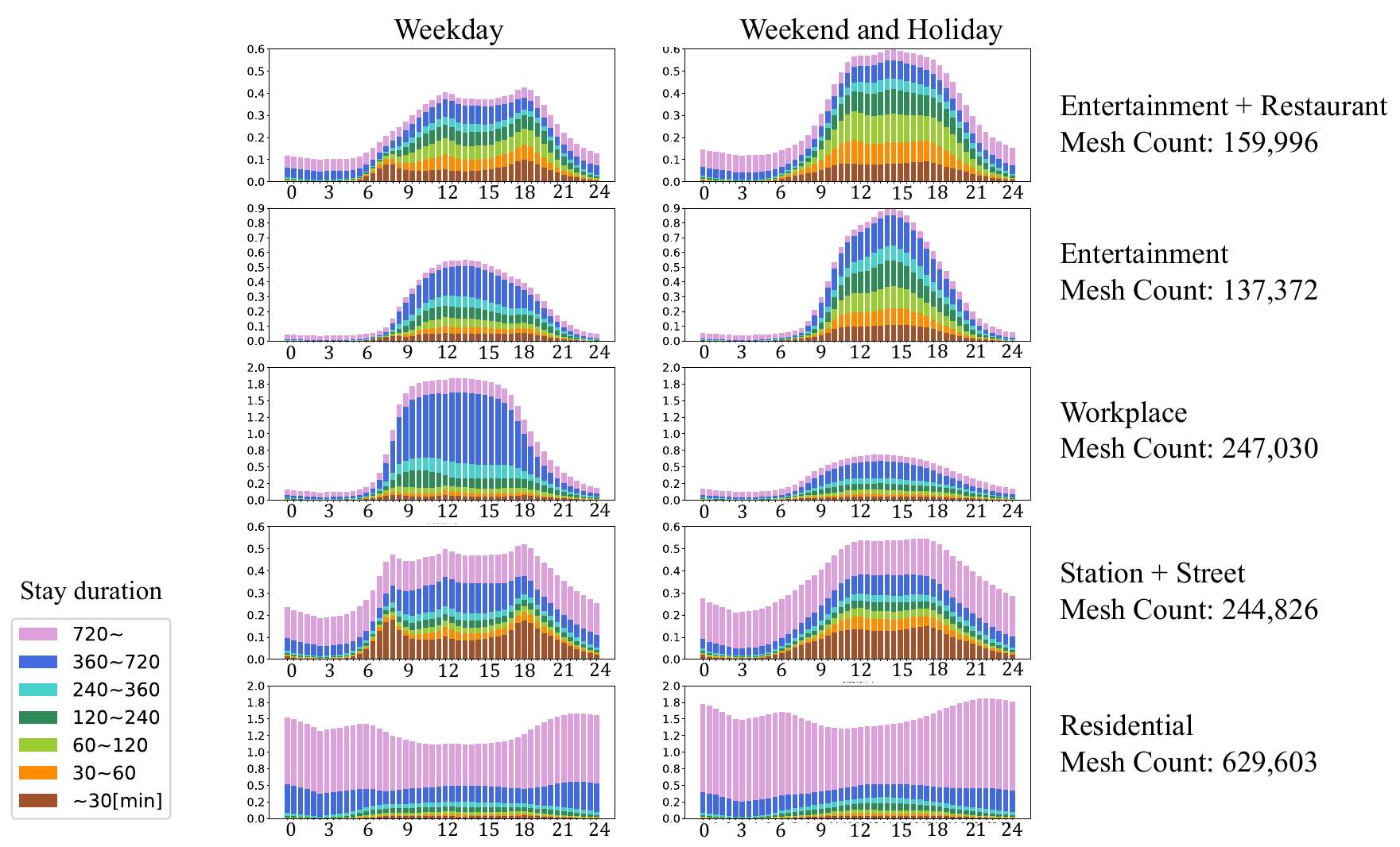}
\caption{Area usage patterns of each 5 cluster, including all cities and periods($D^*$).}
\label{fig:cluster5}
\end{figure}

\section*{Technical Validation}
\subsection*{Effectiveness of cross-city and cross-period analysis using our datasets}
In this section, we demonstrate the utility of our data by practicing a part of cross-city and cross-period area analysis using clustering of area embeddings, thereby showing its effectiveness. Here, we embedded all areas from all cities and periods into the same latent space and used k-means++ for clustering. The number of clusters was set to five, but this number is arbitrary and can be increased for more detailed analysis. Graphs for when the number of clusters was set to 10 and 20 are included in the appendix section. For different periods of data in Nagoya, we treated each period as a distinct area. For example, the areas of Nagoya in January 2021 and February 2021 are considered as different areas. This allows us to analyze how area clusters change over time. The resulting stacked graph is shown in Figure \ref{fig:cluster5}. The horizontal axis represents time, and the vertical axis represents the number of visits per area every 30 minutes, with the color indicating the length of stay. On the right side of the graph, we display labeled interpretations(manually predicted) based on the tendencies of each cluster, along with the count of areas belonging to each cluster. For instance, the first and second clusters, with many short stays on weekends compared to weekdays, can be inferred as areas used like entertainment districts. The third cluster, with many stays over 360 minutes during weekdays, appears to be an office district cluster, while the fourth cluster, with frequent stays under 30 minutes during weekday mornings and evenings, seems to be areas with stations or streets. The fifth cluster, with many stays over 720 minutes, mainly during weekend nights, is a residential area cluster. Thus, areas can be clustered based on their functional differences, and each interpretation of the cluster can be labeled based on the inferred stay tendencies from the graphs.

\begin{figure}[t]
\centering
\includegraphics[width=\linewidth]{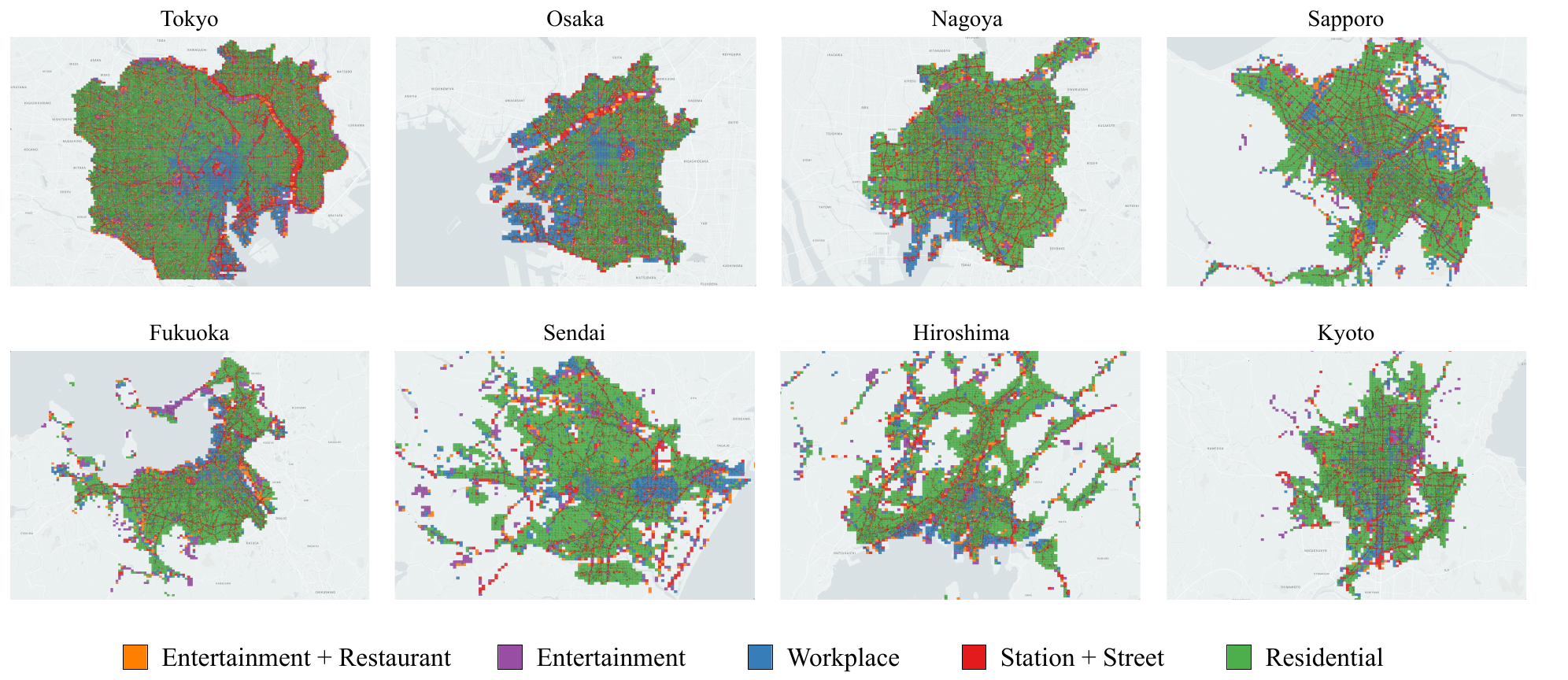}
\caption{Clustering results of UAS in each city.}
\label{fig:map_each_city}
\end{figure}

Next, we visualize the geographical map of these area clusters. We begin by showing the map of area clusters for each city in April 2021 in Figure \ref{fig:map_each_city}. From the figure, we can observe the distribution of areas with distinct usage patterns, such as central train stations and street areas, industrial and office regions in downtown and bay areas, and suburban residential districts. An important point to note here is that despite each area being embedded without sharing location datasets, the anchoring method ensures that each color representing an area function has the same meaning across different cities. Simultaneously, other researchers can embed and analyze areas in other cities using their own data. Thus, our data and method facilitate cross-city analysis of area functions.

\begin{figure}[t]
\centering
\includegraphics[width=\linewidth]{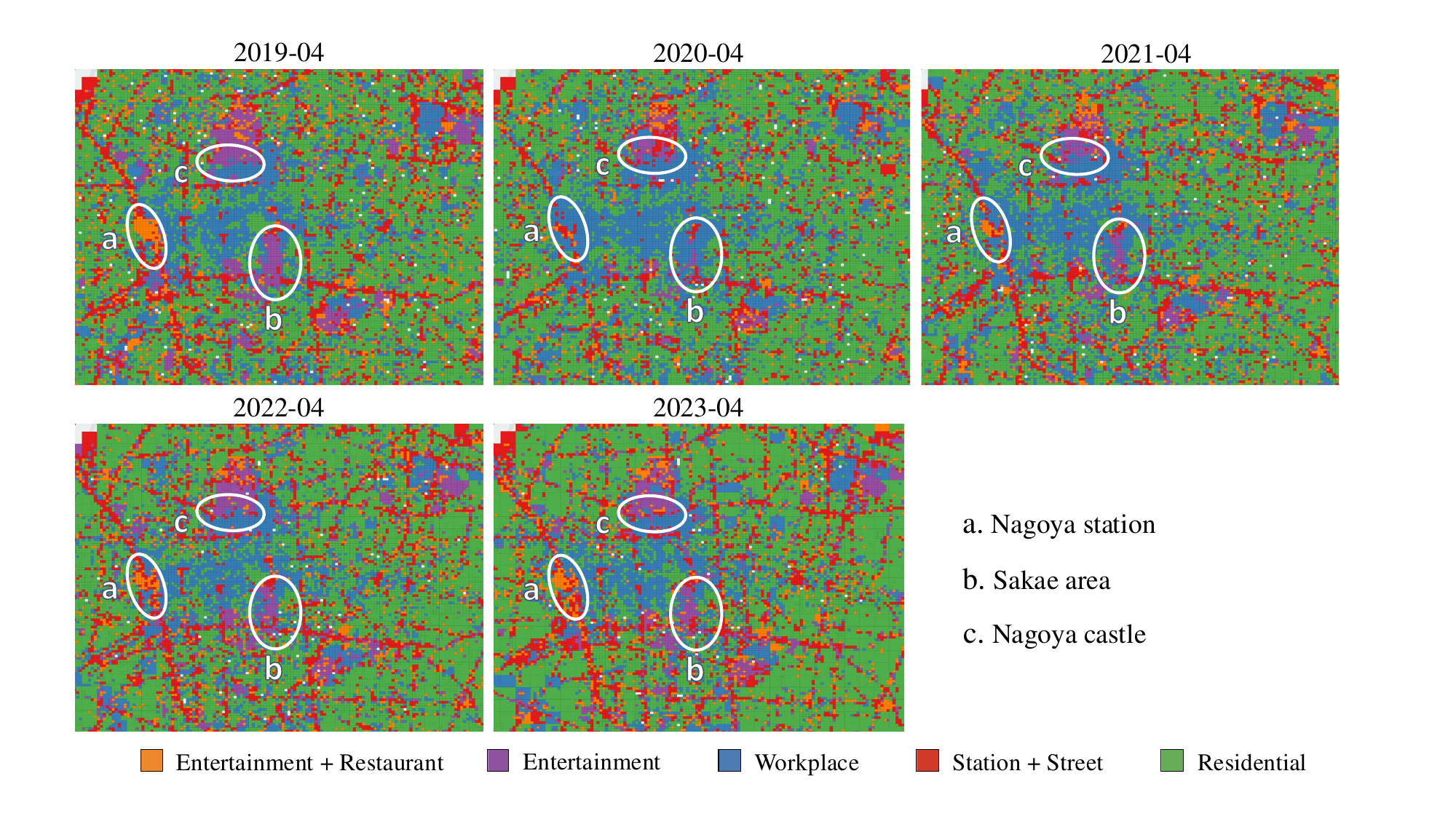}
\caption{Clustering results of UAS for each year. The Entertainment cluster distribution decreased in 2020 due to the COVID-19 pandemic, followed by a recovery trend.}
\label{fig:map_each_year}
\end{figure}

Next, we evaluate cross-period area analysis. We present the distribution of area clusters in the central part of Nagoya for April from 2019 to 2023 in Figure \ref{fig:map_each_year}. Area (a) around Nagoya Station is a region where commercial facilities, restaurants, and offices are located. In 2019, many of these areas belonged to the entertainment and restaurant cluster, but in 2020, many shifted to the workplace cluster. This change reflects the impact of the COVID-19 pandemic in 2020, indicating a decrease in visitors to entertainment and restaurant areas due to self-quarantines. Area (b) is Sakae, a major entertainment area in Nagoya. Here, too, the function of entertainment areas decreased in 2020, but there is a visible recovery of this function in 2021. Area (c) is around Nagoya Castle, a tourist spot, with government buildings such as the city hall to its south. In 2020, this area transitioned to a station and street area, likely due to decreased visitors for sightseeing purposes, while visits for commuting to the workplace area south of Nagoya Castle remained. Annual data, as illustrated here, enable analysis of area changes before and after the COVID-19 pandemic.

\begin{figure}[t]
\centering
\includegraphics[width=\linewidth]{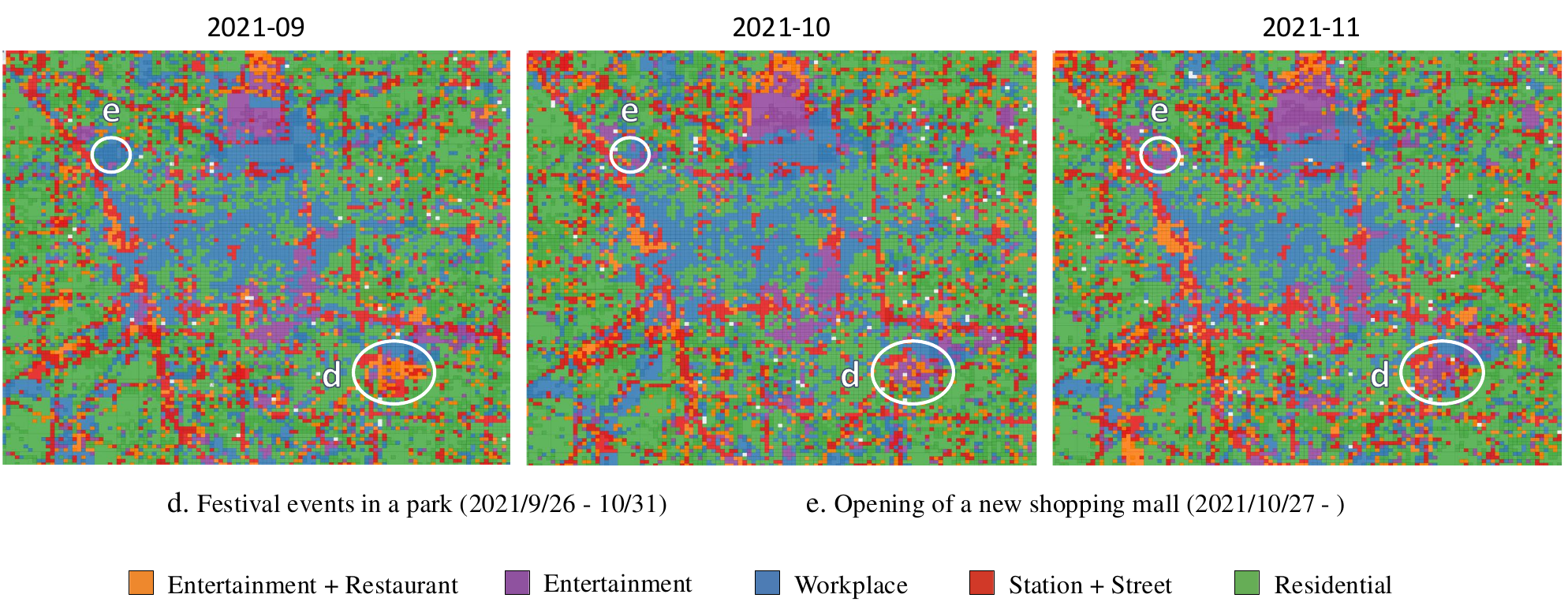}
\caption{Clustering results of UAS for each month. Region d, which hosted a festival, shows changes in trends in October and November. Region e, where a new shopping mall was built, shifted to an Entertainment area in November.}
\label{fig:map_each_month}
\end{figure}

Finally, we visualize the monthly changes of areas within the same region. We have chosen and visualized the three months from September to November 2021 from the entire year, as shown in Figure\ref{fig:map_each_month}. Area (d) includes a park, where a festival event was held from September 26 to October 31, 2021. The impact of this event, coupled with the lifting of the state of emergency from August 27 to September 30, 2021, can be seen in the changing functions of the area. In area (e), a large shopping mall opened on October 27, 2021. This led to the rise of an entertainment area in a region that previously only had workplaces and residential areas. Thus, using annual and monthly data, it is possible to analyze changes in the functions of the same area over time.

\subsection*{Validation of anchoring method}
Next, we validate the effectiveness of the anchoring method. The key aspect of the anchoring method is the ability to embed areas in different datasets into the common embedding space without sharing datasets. In traditional methods without anchoring technique, even if the exact same dataset is embedded, the first and second embeddings are positioned far apart, failing to share the embedding space. In this method, areas in each dataset are embedded into a common embedding space $E^+$ through the anchor data $D_a$. To evaluate whether they are appropriately embedded in the common space, we embed $D^+$ twice, comparing the results with and without the anchoring method. If the anchoring is not functioning correctly, the first and second embeddings will be significantly distant from each other in the latent space. We then define Misalignment, as shown in the following equation, as the discrepancy between the obtained embedding $E$ and $E^+$

\begin{equation}
   M\!isalignment(E, E^+) = \frac{1}{|M|} \sum\limits_{m\in M} D\!istance(e_m, e^{+}_m)
   \label{misalignment}
\end{equation}

\begin{figure}[t]
\centering
\includegraphics[width=\linewidth]{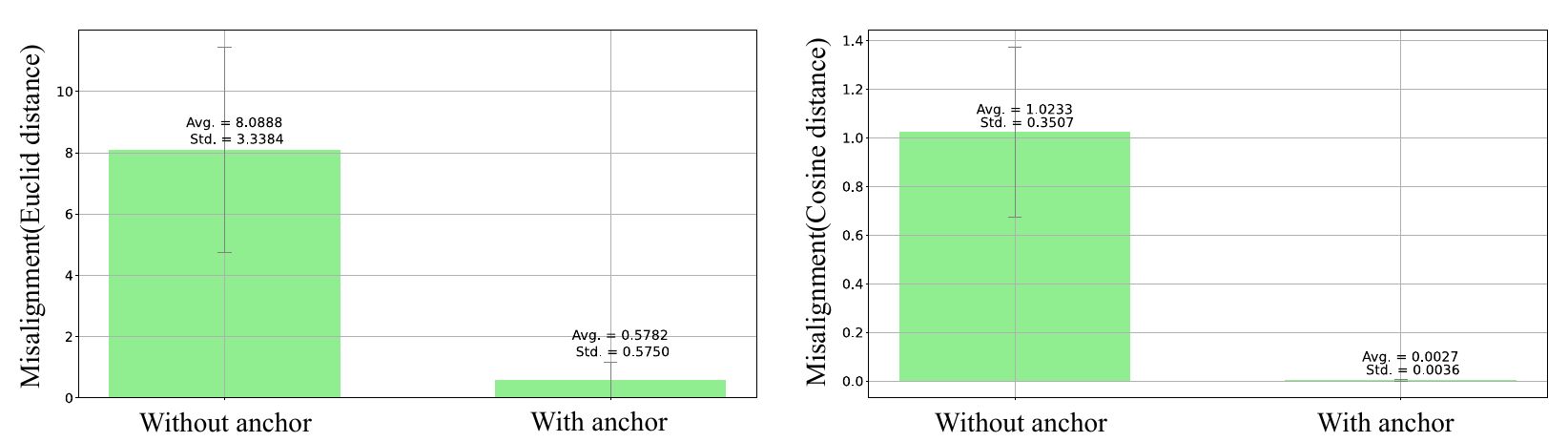}
\caption{Comparison of $Misalignment$ values with and without anchor learning.}
\label{fig:misalignment}
\end{figure}

$M$ represents the set of all areas in the dataset $D^*$, with each area denoted as $m$ and the total number of areas in $M$ as $|M|$. $e_m$ represents the retrained embedding corresponding to $m$, whereas $e_m^+$ represents the embedding for $m$ in $E^+$. A smaller $Misalignment(E, E^+)$ indicates a greater success in embedding each area into the same latent space as $E^+$ in the retraining. $Distance$ is a distance function for which both Euclidean distance and Cosine distance are used. Cosine distance is calculated as 1 minus the cosine similarity. Here, we compared the original Area2Vec without anchoring and Area2Vec with anchoring. Results are shown in Figure \ref{fig:misalignment}. From these results, it is evident that the original method yields different embeddings for independent learning, indicating that it does not embed in a common latent space. In contrast, our proposed anchoring method resulted in a smaller variance in embeddings for the same data, demonstrating the ability to embed areas into the same representation space.

Simultaneously, it is necessary to evaluate whether the ability in capturing area usage patterns has been compromised by anchoring. This concern arises because the anchoring method may distort the update of weights during learning via anchors, potentially preventing the acquisition of optimal area embeddings for the training data. In particular, area data with usage patterns not included in anchor data may not be representable in the embedding space of OpenUAS. To validate this, we conducted the following experiments for each dataset comprising $D^*$. 
\begin{enumerate}
    \item Construct anchor data using $D^*$ excluding the target dataset (such as $D_{tokyo}$).
    \item Apply this anchor data to train the target dataset while anchoring it.
    \item Compare the embeddings learned from the target dataset individually with those trained in step 2.
\end{enumerate}
If the performance using anchoring significantly declines compared to when learning individually, it may indicate that the anchor data does not fully capture the latent patterns of areas within the target data, failing to anchor them appropriately in the ideal space.

Although area embedding is an unsupervised learning process, poses challenges for direct performance validation. To address this, we introduce the metric $ApproximationLoss$. Area embedding serves as a method of dimension reduction. During the training process, the product of the embedding $e_m$ for area m and the output matrix $W'$ is optimized to better approximate the frequency $f_m$ of the discretized stay feature for each area. Here, $f_m$ is a 168-dimensional vector representing the frequency of the one-hot for discretized stay features. Essentially, the performance of area embedding improves as the product of $e_m$ and $W'$ more closely approximates $f_m$. This discrepancy is quantified by $ApproximationLoss$. The corresponding formula is presented below.

\begin{equation}
   A\!pproximationLoss = \frac{1}{|M|} \sum\limits_{m\in M} (1 - Cos(\hat{f_m}, f_m))
   \label{rloss}
\end{equation}

\begin{equation}
   \hat{f_m} = S\!oftmax(e_mW')
   \label{pred_feature}
\end{equation}

\begin{figure}[t]
\centering
\includegraphics[width=\linewidth]{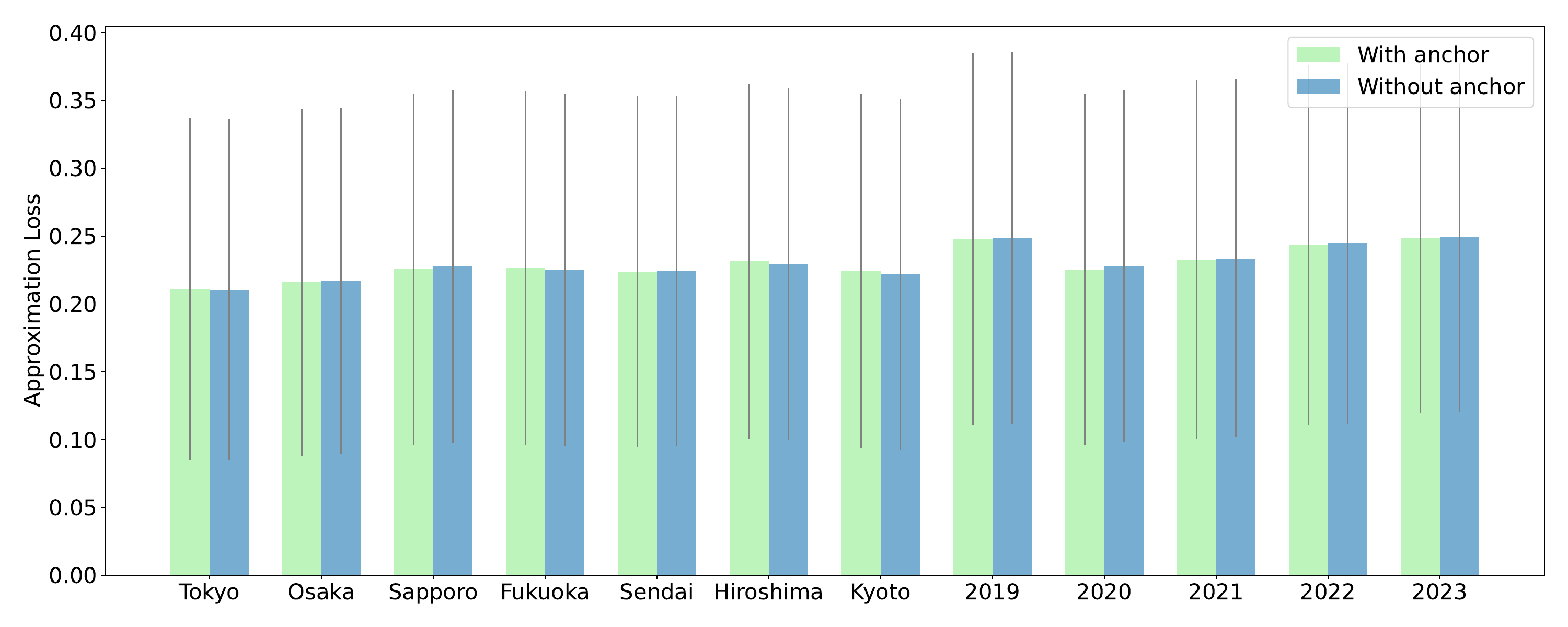}
\caption{Comparison of $ApproximationLoss$ values with and without anchor-based learning. While the loss is optimal when learning freely without anchors, there is no significant difference when learning with anchors.}
\label{fig:approximation_loss}
\end{figure}

Here, $f_m$ represents the true frequency of discretized stay features in area $m$, and $\hat{f_m}$ represents the approximate frequency calculated as the product of $e_m$ and $W'$. We conducted the previously described experiments on $D_{tokyo}$ to $D_{2023}$, which make up $D^*$, calculating and comparing the $ApproximationLoss$ learned using the original Area2Vec(without anchoring) and the anchoring method. The results are shown in Figure \ref{fig:approximation_loss}. The results reveal that there is no significant difference in the $ApproximationLoss$ values before and after the application of the anchoring method. However, it is also noted that the $ApproximationLoss$ values differ by a few points regardless of the application of the anchoring method, with the differences being especially prominent across different years. This variation is believed to be caused by the differences in latent trends of areas across the datasets. For instance, the year 2019 shows worse metrics compared to other datasets, which could be attributed to the greater diversity in area characteristics, making it more challenging to represent them with fewer embedding dimensions. The variations in embedding accuracy due to differences in cities and years suggest that further evaluation is needed using data from cities outside Japan. Nonetheless, the minor deterioration in embedding accuracy before and after applying anchoring indicates that the method is effective.

\subsection*{Anonimity}
Finally, we evaluate our area embedding datasets from the perspective of anonymity. According to the GDPR, data such as GPS movement data should be sufficiently anonymized to make the re-identification of individuals impossible\cite{GDPR}. Our data is anonymized and abstracted in two steps.

In the first step, stay features in the location data are aggregated by mesh. This process statistically represents location data for each mesh, forming an m $\times$ q dimensional matrix (where m is the number of meshes and q is the number of discretized stay features). In other words, by aggregating location data tied to individual user IDs by mesh, we eliminate personal data. At the same time, by focusing only on meshes visited by more than 10 people, we reduce the possibility of these statistics being linked to specific individuals or households. However, even with this anonymization, directly releasing this statistical information could pose a risk, as it would lead to the direct disclosure of the stay features for each mesh. For instance, if it becomes clear that there are no users in a certain mesh at a certain time, this information could be exploited for crimes like burglary. Therefore, it is difficult to release this statistical information directly.

\begin{figure}[t]
\centering
\includegraphics[width=\linewidth]{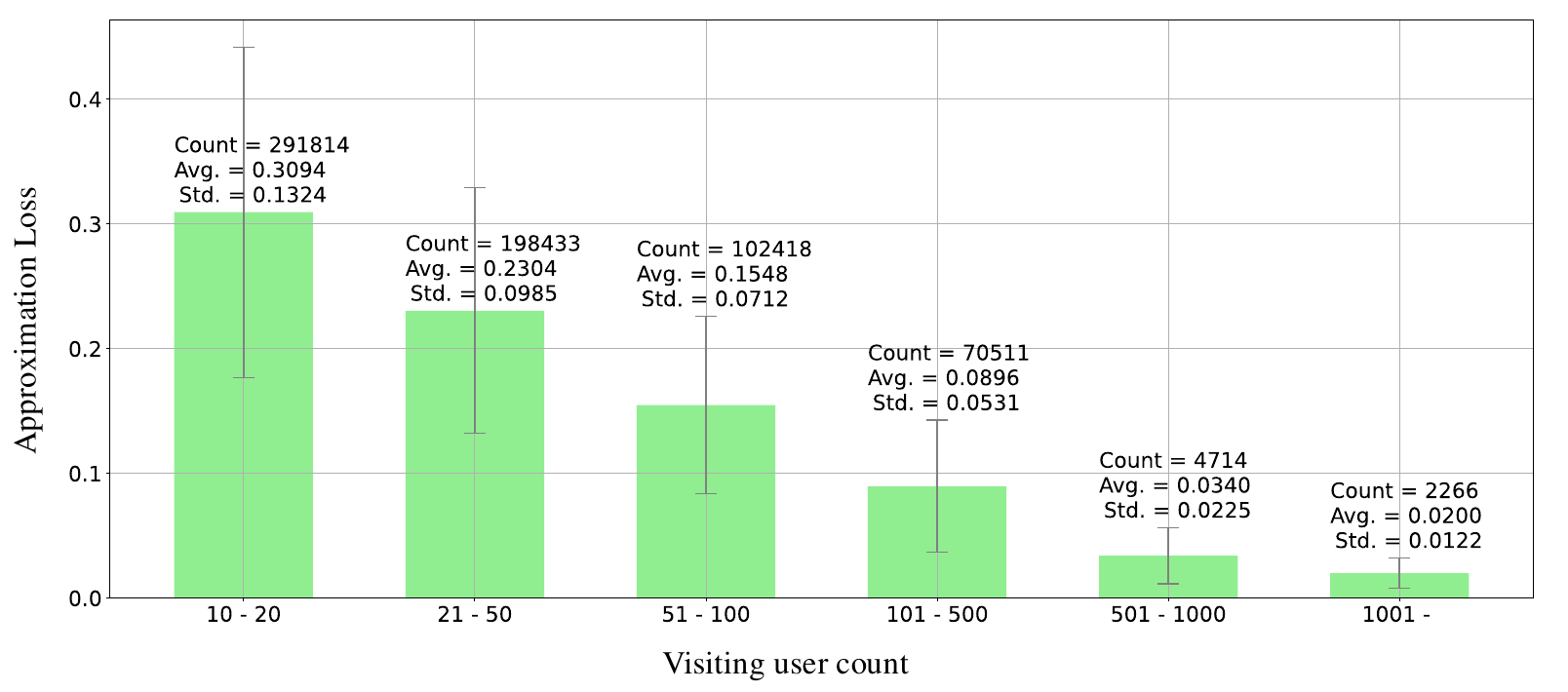}
\caption{Relationship between user counts and  $ApproximationLoss$.}
\label{fig:anonymity}
\end{figure}

In the second step, these statistical vectors are embedded into an 8-dimensional latent space, further abstracting the information. As shown in the Technical Validation section, the degree of match between the approximated and actual statistical vectors for each mesh is calculated by $ApproximationLoss$. Therefore, data users can only know the approximate values of the statistics for each mesh. Moreover, $ApproximationLoss$ tends to be higher for meshes with more user counts and lower for those with fewer. Figure \ref{fig:anonymity} illustrates the relationship between user count per mesh and $ApproximationLoss$. From Figure \ref{fig:anonymity}, it is apparent that in meshes with 10-20 user counts, the values are higher and decrease as the user count increases. This result suggests it is particularly difficult to infer related statistical information in meshes with fewer users. Therefore, our data does not contain personal data and only provides approximate values for any statistical information that could indirectly estimate personal data. From these aspects, it can be concluded that our data is sufficiently anonymized.

\section*{Usage Notes}
Here, we introduce several applications that effectively utilize area embedding and anchor data. The embeddings of Japanese cities use both 50m and 250m meshes, as described in the Methods section. However, for the 50m meshes with fewer users, we complement them with the embeddings of the encompassing 250m meshes. 

\begin{description}
\item[Embedding with Anchoring]If users have their own location data, they can train Area2Vec and acquire embeddings for each area. In addition to unique area embeddings from their data, our developed anchoring method allows comparison with OpenUAS embeddings.
\item[Area Clustering]This involves clustering a group of target area embeddings into any number of clusters, forming clusters of areas with similar functions. This allows for the classification and analysis of urban sections based on their functions and usage. Our code includes the tool for clustering area embeddings into any number of clusters and the tool for geographical visualization.
\item[Similar Area Searching]Users can select any area and search for similar areas. Adjusting the similarity threshold enables visualization of multiple neighboring areas.
\item[Approximation of Area Trends]As explained in the Technical Validation, area trends can be predicted through the approximation of discretized stay statistics. While not perfectly matching actual statistics, this method can reproduce stay trend graphs as shown in Figure \ref{fig:cluster5}, which is effective for area analysis.
\end{description}

Additionally, while not covered here, area embedding has potential applications in downstream machine learning tasks like predicting human movement destinations and recommendations\cite{poi_recommend1, poi_recommend2, areavec3, city2city, hier_emb, haru_humob}.

\section*{Code availability}
The code for area analysis is available on the GitHub repository(\url{https://github.com/UCLabNU/OpenUAS/}). By downloading our data and placing it in the directory, it is possible to visualize and analyze the OpenUAS embeddings. Since the raw location dataset for embedding cannot be shared due to privacy and confidentiality reasons, synthetic sample data is provided within the repository. By replacing these with your own dataset, it is possible to embed and compare within the same space as the OpenUAS embeddings.

\bibliography{sample}

\section*{Acknowledgements}
 This research was supported in part by JST CREST (JPMJCR21F2, JPMJCR22M4), ACT-X(JPMJAX24M8), NICT (222C01, 22609), NEDO (JPNP23003), CSTI SIP3(JPJ012495), and JSPS KAKENHI (22H03580, 22K18422). We would like to express our gratitude to Blogwatcher, Inc.\cite{blogwatcher} for their cooperation in providing the data.

\section*{Author contributions statement}
N.T. and K.S. conducted the experiments. N.T. prepared the first draft of the manuscript. S.K., K.U., T.Y., and N.K. supervised the project. All authors reviewed and approved the final manuscript for submission.

\section*{Competing interests}
The authors declare that they have no competing interests.

\section*{Appendix}

\subsection*{Number of anchor and anchor dataset size}

\begin{figure}[t]
\centering
\includegraphics[width=\linewidth]{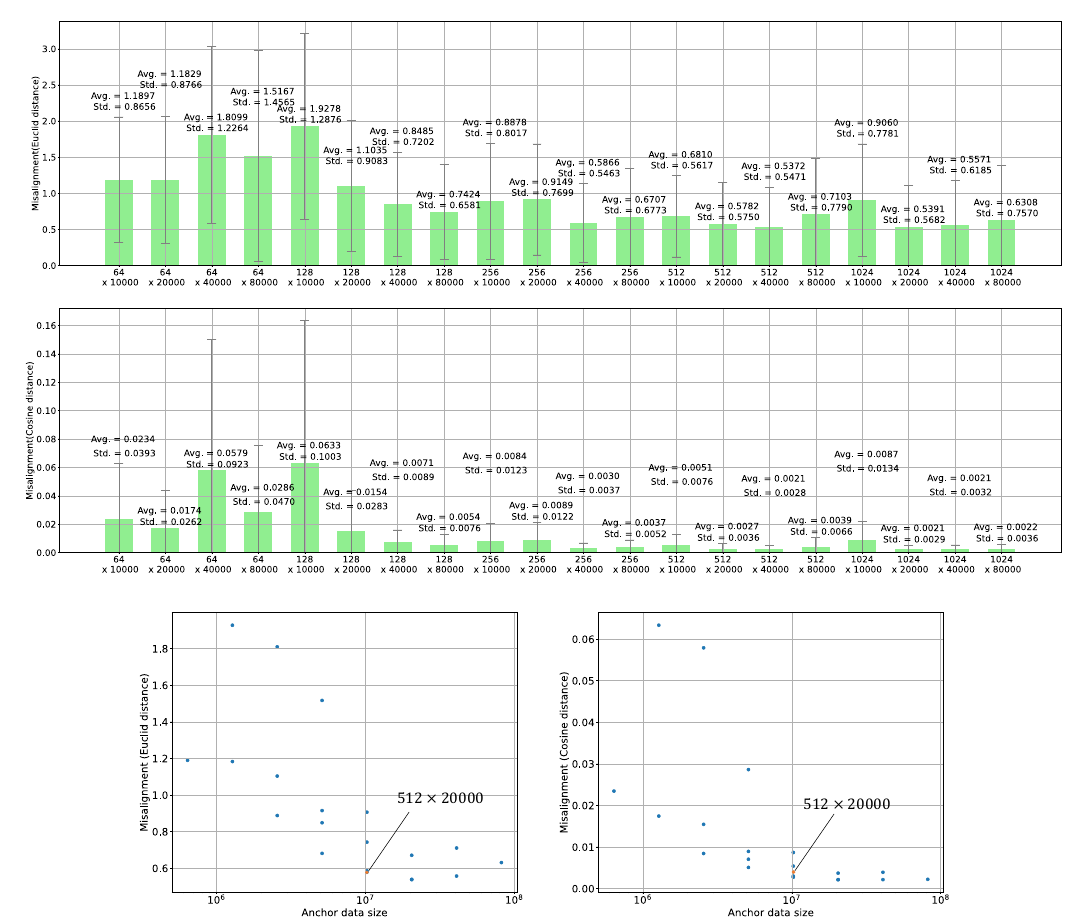}
\caption{Comparison of number of anchors and anchor dataset size.}
\label{fig:anchor_size}
\end{figure}

To select the number of anchors and the size of data each anchor holds, we compared several variations. In this paper, we experimented with 64 to 1024 anchors and data sizes ranging from 10,000 to 80,000, comparing the $Misalignment$ for each variation. Figure \ref{fig:anchor_size} shows the results. The lower graph is a scatter plot with the data size of each variation on the x-axis. The results indicate that $Misalignment$ decreases as the size of the anchor data increases. However, larger anchor data sizes increase the volume of training data, leading to longer training times and greater memory usage. Considering these costs, we used anchor data of size $512\times20,000$ in the OpenUAS dataset. 

\subsection*{Types of anchor weight functions}
\begin{figure}[t]
\centering
\includegraphics[width=\linewidth]{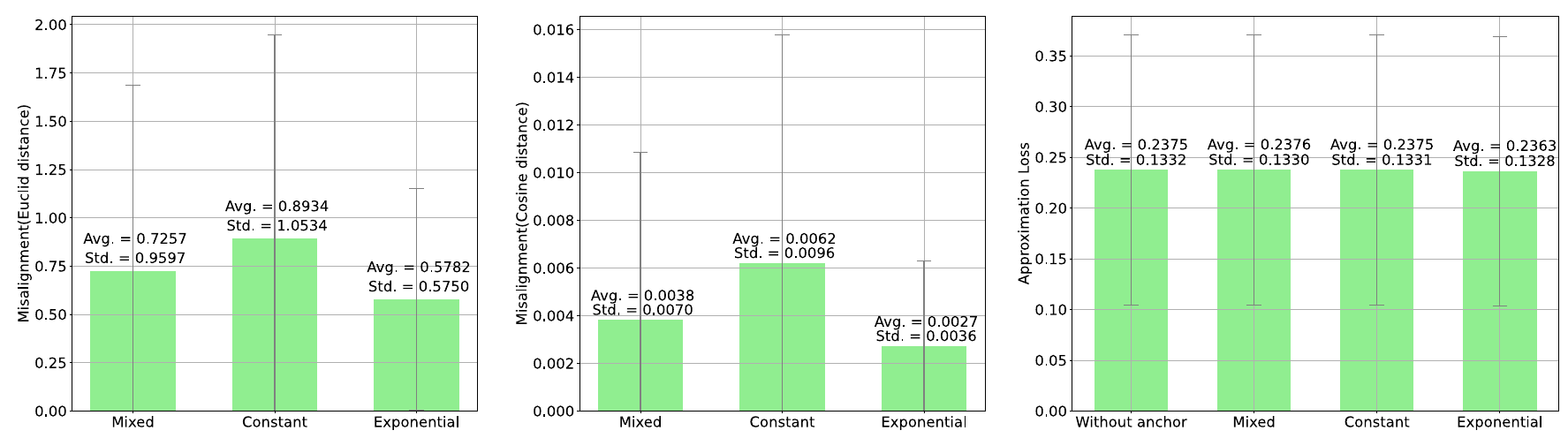}
\caption{Comparison of types of anchor weight functions.}
\label{fig:anchor_weight}
\end{figure}
We adjusted the ratio of loss backpropagation for data and anchors according to formulas \ref{anchor_loss}, \ref{anchor_power} during anchoring. In Figure \ref{fig:anchor_weight}, we examine the effects of various anchor weight functions. `Mixed' simply combines the anchor data with the training data, treating the anchors in the same way as other areas in the data without any weighting. `Constant' fixes the anchor power at $\alpha = 0.3$ from the beginning to the end of the training, regardless of the epoch. `Exponential' decreases anchor power exponentially, as shown in the formula \ref{anchor_power}. According to the results, the `Exponential' method performs better in all metrics. Additionally, `Mixed' has better metrics than `Constant', which may be due to the relatively large size of anchor data ($512\times20,000$) in compared to the training data in our experiments. Conversely, this suggests that if the anchor data size is small compared to the training data, simply mixing the anchor data with the training data might lead to the anchor data being ignored during learning.

\subsection*{Value of $\alpha$}
\begin{figure}[t]
\centering
\includegraphics[width=\linewidth]{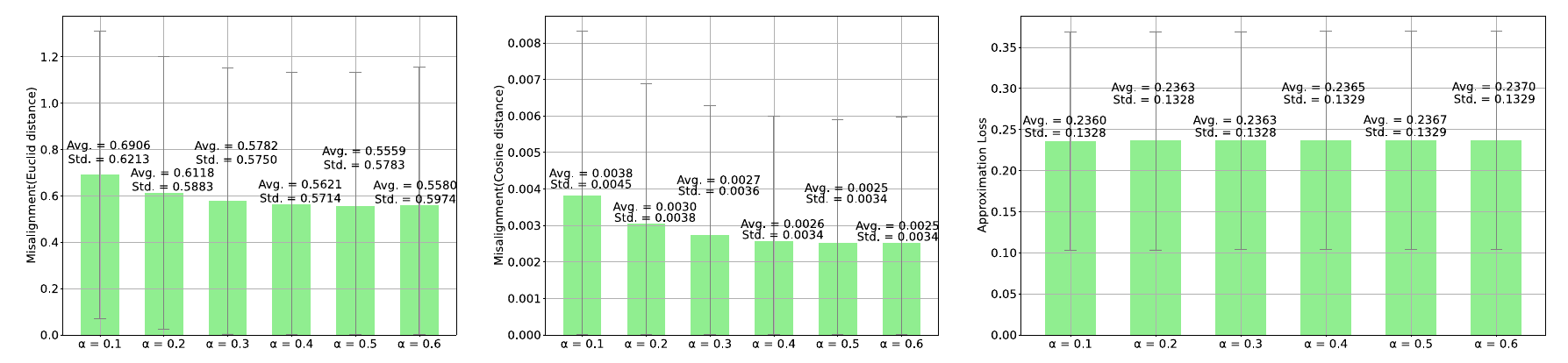}
\caption{Comparison of value of $\alpha$(in Equation \ref{anchor_power}).}
\label{fig:alpha}
\end{figure}
Finally, we consider the final value of anchor power, denoted as $\alpha$ in Equation \ref{anchor_power}. In Figure \ref{fig:alpha}, we compared values ranging from $0.1$ to $0.6$. For $Misalignment$, the metrics improve as $\alpha$ increases, indicating that a higher anchor power during training leads to a slightly smaller $Misalignment$. Conversely, for $ApproximationLoss$, the metric is somewhat higher with a larger anchor power. To balance these two indicators, we chose $\alpha = 0.3$.

\end{document}